# Large-Scale Feature Learning With Spike-and-Slab Sparse Coding

Ian J. Goodfellow  goodfeli.@iro.umontreal.ca
Aaron Courville  Aaron.Courville@umontreal.ca
Yoshua Bengio  Yoshua.Bengio@umontreal.ca
DIRO, Université de Montréal, Montréal, Québec, Canada

## Abstract

We consider the problem of object recognition with a large number of classes. In order to overcome the low amount of labeled examples available in this setting, we introduce a new feature learning and extraction procedure based on a factor model we call *spike-and-slab sparse coding* (S3C). Prior work on S3C has not prioritized the ability to exploit parallel architectures and scale S3C to the enormous problem sizes needed for object recognition. We present a novel inference procedure for appropriate for use with GPUs which allows us to dramatically increase both the training set size and the amount of latent factors that S3C may be trained with. We demonstrate that this approach improves upon the supervised learning capabilities of both sparse coding and the spike-and-slab Restricted Boltzmann Machine (ssRBM) on the CIFAR-10 dataset. We use the CIFAR-100 dataset to demonstrate that our method scales to large numbers of classes better than previous methods. Finally, we use our method to win the NIPS 2011 Workshop on Challenges In Learning Hierarchical Models' Transfer Learning Challenge.

## 1. Introduction

We consider here the problem of unsupervised feature discovery for supervised learning. In supervised learning, one is given a set of examples $V = \{v^{(1)}, \ldots, v^{(m)}\}$ and associated labels $\{y^{(1)}, \ldots, y^{(m)}\}$. The goal is to learn a model $p(y \mid v)$ so that new labels can be predicted from new unlabeled examples $v$.

The idea behind unsupervised feature discovery is that the final learning problem can become much easier if the problem is represented in the right way. By learning the structure of $V$ we can discover a feature mapping $\phi(v)$ that can be used to preprocess the data prior to running a standard supervised learning algorithm, such as an SVM.

There has been a great deal of recent interest in investigating different unsupervised learning schemes to train $\phi$ from $V$. In particular, the goal of deep learning (Bengio, 2009) is to learn a function $\phi$ that consists of many layers of processing, each of which receives the previous layers as input and incrementally disentangles the factors of variation in the data. Deep learning systems are usually created by composing together several shallow unsupervised feature learners. Examples of shallow models applied to feature discovery include sparse coding (Raina et al., 2007), restricted Boltzmann machines (RBMs) (Hinton et al., 2006; Courville et al., 2011b), various autoencoder-based models (Bengio et al., 2007), and hybrids of autoencoders and sparse coding (Kavukcuoglu et al., 2010). In the context of probabilistic generative models, such as the RBM, $\phi(v)$ is typically taken to be the conditional expectation of the latent variables, and the process of learning $\phi$ consists simply of fitting the generative model to $V$.

Single-layer convolutional models based on simple feature extractors currently achieve state-of-the-art performance on the CIFAR-10 object recognition dataset (Coates and Ng, 2011; Jia and Huang, 2011). It is known that the best models for the detection layer of the convolutional model do not perform well when fewer labeled examples are available (Coates and Ng, 2011). In particular, sparse coding outperforms a simple thresholded linear feature extractor when the number of labeled examples decreases. Our objective is to further improve performance when the number of labeled examples is low by introducing a new feature extraction procedure based on spike-and-slab sparse coding. We hypothesize that these features have a stronger regularizing effect than sparse coding features. Their superior performance with low numbers of labeled examples allows us to improve performance on datasets with high numbers of classes and low numbers of labeled examples.

In this paper we overcome two major scaling challenges. First we scale inference in the spike-and-slab coding model to work for the large problem sizes required for object recognition. We then use the enhanced regularization properties of spike-and-slab sparse coding to scale object recognition techniques to work with large numbers of classes and small amounts of labeled data.

## 2. The Spike-and-Slab Sparse Coding model

The Spike-and-Slab Sparse Coding (S3C) model consists of latent binary *spike* variables $h \in \{0,1\}^N$, latent real-valued *slab* variables $s \in \mathbb{R}^N$, and real-valued $D$-dimensional visible vector $v \in \mathbb{R}^D$ generated according to this process:

$\forall i \in \{1, \ldots, N\}, d \in \{1, \ldots, D\},$

$$\begin{aligned} p(h_i = 1) &= \sigma(b_i) \\ p(s_i \mid h_i) &= \mathcal{N}(s_i \mid h_i \mu_i, \alpha_{ii}^{-1}) \\ p(v_d \mid s, h) &= \mathcal{N}(v_d \mid W_{d:}(h \circ s), \beta_{dd}^{-1}) \end{aligned} \qquad (1)$$

where $\sigma$ is the logistic sigmoid function, $b$ is a set of biases on the spike variables, $\mu$ and $W$ govern the linear dependence of $s$ on $h$ and $v$ on $s$ respectively, $\alpha$ and $\beta$ are diagonal precision matrices of their respective conditionals, and $h \circ s$ denotes the element-wise product of $h$ and $s$.

To avoid overparameterizing the distribution, we constrain the columns of $W$ to have unit norm, as in sparse coding. We restrict $\alpha$ to be a diagonal matrix and $\beta$ to be a diagonal matrix or a scalar. We refer to the variables $h_i$ and $s_i$ as jointly defining the $i^{\text{th}}$ hidden unit, so that there are a total of $N$ rather than $2N$ hidden units. The state of a hidden unit is best understood as $h_i s_i$, that is, the spike variables gate the slab variables.

In the subsequent sections we motivate our use of S3C as a feature discovery algorithm by describing how this model occupies a middle ground between sparse coding and the spike-and-slab Restricted Boltzmann Machine (ssRBM). The S3C model avoids many disadvantages that the ssRBM and sparse coding have when applied as feature discovery algorithms.

### 2.1. Comparison to sparse coding

Sparse coding has been widely used to discover features for classification (Raina et al., 2007). Recently Coates and Ng (2011) showed that this approach achieves excellent performance on the CIFAR-10 object recognition dataset.

Sparse coding (Olshausen and Field, 1997) describes a class of generative models where the observed data $v$ is normally distributed given a set of continuous latent variables $s$ and a dictionary matrix $W$: $v \sim \mathcal{N}(Ws, \sigma \mathbb{I})$. Sparse coding places a factorial prior on $s$ such as a Cauchy or Laplace distribution, chosen to encourage the mode of the posterior $p(s \mid v)$ to be sparse. One can derive the S3C model from sparse coding by replacing the factorial Cauchy or Laplace prior with a spike-and-slab prior.

One drawback of sparse coding is that the latent variables are not merely encouraged to be sparse; they are encouraged to remain close to 0, even when they are active. This kind of regularization is not necessarily undesirable, but in the case of simple but popular priors such as the Laplace prior (corresponding to an $L_1$ penalty on the latent variables $s$), the degree of regularization on active units is confounded with the degree of sparsity. There is little reason to believe that in realistic settings, these two types of complexity control should be so tightly bound together. The S3C model avoids this issue by controlling the sparsity of units via the $b$ parameter that determines how likely each spike unit is to be active, while separately controlling the magnitude of active units via the $\mu$ and $\alpha$ parameters that govern the distribution over $s$. Sparse coding has no parameter analogous to $\mu$ and cannot control these aspects of the posterior independently.

Another drawback of sparse coding is that the factors are not actually sparse in the generative distribution. Indeed, each factor is zero with probability zero. The features extracted by sparse coding are only sparse because they are obtained via MAP inference. In the S3C model, the spike variables ensure that each factor is zero with non-zero probability in the generative distribution. Since this places a greater restriction on the code variables, we hypothesize that S3C features provide more of a regularizing effect when solving classification problems.

Sparse coding is also difficult to integrate into a deep generative model of data such as natural images. While Yu et al. (2011) and Zeiler et al. (2011) have recently shown some success at learning hierarchical sparse coding, our goal for our future work is to integrate the feature extraction scheme into a proven generative model framework such as the deep Boltzmann machine (Salakhutdinov and Hinton, 2009). Such models with their combination of feed-forward and feed-back connections during inference can learn a much richer description of the data than simple stacked feed-forward models. We expect that being able to extract such complicated structure during unsupervised learning on a large number of unlabeled examples will yield even better performance with high numbers of classes and low numbers of labels per class than a feed-forward architecture. Existing inference schemes known to work well in the DBM-type setting are all either sample-based or are based on variational approximations to the model posteriors, while sparse coding schemes typically employ MAP inference. Our use of variational inference makes the S3C framework well-suited to integrate into the known successful strategies for learning and inference in DBM models. It is not obvious how one can employ a variational inference strategy to standard sparse coding with the goal of achieving sparse feature encoding.

### 2.2. Comparison to Restricted Boltzmann Machines

The S3C model also resembles another class of models commonly used for feature discovery: the RBM. An RBM (Smolensky, 1986) is a model defined through an energy function that describes the interactions between the observed data variables and a set of latent variables. It is possible to interpret the S3C as an energy-based model, by rearranging $p(v, s, h)$ to take the form $\exp\{-E(v, s, h)\}/Z$, with the following energy function:

$$E(v, s, h) = \frac{1}{2}\left(v - \sum_i W_i s_i h_i\right)^T \beta \left(v - \sum_i W_i s_i h_i\right)$$
$$+ \frac{1}{2}\sum_{i=1}^N \alpha_i (s_i - \mu_i h_i)^2 - \sum_{i=1}^N b_i h_i, \quad (2)$$

The ssRBM model family is a good starting point for S3C because it has demonstrated both reasonable performance as a feature discovery scheme and remarkable performance as a generative model (Courville et al., 2011a). Within the ssRBM family, S3C's closest relative is a variant of the $\mu$-ssRBM, defined by the following energy function:

$$E(v, s, h) = -\sum_{i=1}^N v^T \beta W_i s_i h_i + \frac{1}{2} v^T \beta v$$
$$+ \frac{1}{2}\sum_{i=1}^N \alpha_i (s_i - \mu_i h_i)^2 - \sum_{i=1}^N b_i h_i, \quad (3)$$

where the variables and parameters are defined identically to the S3C. Comparison of equations 2 and 3 reveals that the simple addition of a latent factor interaction term $\frac{1}{2}(h \circ s)^T W^T \beta W (h \circ s)$ to the ssRBM energy function turns the ssRBM into the S3C model. With the inclusion of this term S3C moves from an undirected ssRBM model to the directed graphical model described in equation (1). One can think of this term as designed to cancel the interactions in the RBM's marginal $p(h, s)$ that make the RBM's partition function intractable. This change from undirected modeling to directed modeling has three important effects, that we describe in the following paragraphs:

**The effect on the partition function:** The most immediate consequence of the transition to directed modeling is that the partition function becomes tractable. This changes the nature of learning algorithms that can be applied to the model, since most of the difficulty in training an RBM comes from estimating the gradient of the log partition function. The partition function of S3C is also guaranteed to exist for all possible settings of the model parameters, which is not true of the ssRBM.

**The effect on the posterior:** RBMs have a factorial posterior, but S3C and sparse coding have a complicated posterior due to the "explaining away" effect. This means that for RBMs, features defined by similar basis functions will have similar activations, while in directed models, similar features will compete so that only the most relevant feature will remain active. As shown by Coates and Ng (2011), the sparse Gaussian RBM is not a very good feature extractor – the set of basis functions $W$ learned by the RBM actually work better for supervised learning when these parameters are plugged into a sparse coding model than when the RBM itself is used for feature extraction. We think this is due to the factorial posterior. In the vastly overcomplete setting, being able to selectively activate a small set of features that cooperate to explain the input likely provides S3C a major advantage in discriminative capability.

**The effect on the prior:** The addition of the interaction term causes S3C to have a factorial prior. This probably makes it a poor generative model, but this is not a problem for the purpose of feature discovery.

### 3. Other Related work

The notion of a spike-and-slab prior was established in statistics by Mitchell and Beauchamp (1988). Outside the context of unsupervised feature discovery for supervised learning, the basic form of the S3C model (i.e. a spike-and-slab latent factor model) has ap-peared a number of times in different domains (Lücke and Sheikh, 2011; Garrigues and Olshausen, 2008; Mohamed et al., 2011; Zhou et al., 2009; Titsias and Lázaro-Gredilla, 2011). In most work, the model varies slightly from S3C. For example, Titsias and Lázaro-Gredilla (2011) share a single spike activation probability parameter across all spike variables. Lücke and Sheikh (2011) use exactly the S3C model, but use intractable exact inference. To this literature, we contribute an approximate inference scheme that scales to the kinds of object classifications tasks that we consider. We outline this inference scheme next.

## 4. Variational EM for S3C

Having explained why S3C is a powerful model for unsupervised feature discovery we turn to the problem of how to perform learning and inference in this model. Because computing the exact posterior distribution is intractable, we derive an efficient and effective inference mechanism and a variational EM learning algorithm.

We turn to variational EM (Saul and Jordan, 1996) because this algorithm is well-suited for models with latent variables whose posterior is intractable. It works by maximizing a variational lower bound on the log-likelihood called the energy functional (Neal and Hinton, 1999). More specifically, it is a variant of the EM algorithm with the modification that in the E-step, we compute a variational approximation to the posterior rather than the posterior itself. While our model admits a closed-form solution to the M-step, we found that online learning with small gradient steps on the M-step objective worked better in practice. We therefore focus our presentation on the E-step, given in Algorithm 1.

The goal of the variational E-step is to maximize the energy functional with respect to a distribution $Q$ over the unobserved variables. We can do this by selecting the $Q$ that minimizes the Kullback–Leibler divergence:

$$\mathcal{D}_{KL}(Q(h,s)\|P(h,s|v)) \qquad (4)$$

where $Q(h, s)$ is drawn from a restricted family of distributions. This family can be chosen to ensure that $Q$ is tractable.

Our E-step can be seen as analogous to the encoding step of the sparse coding algorithm. The key difference is that while sparse coding approximates the true posterior with a MAP point estimate of the latent variables, we approximate the true posterior with the distribution $Q$.

We use the family $Q(h, s) = \Pi_i Q(h_i, s_i)$. This is a

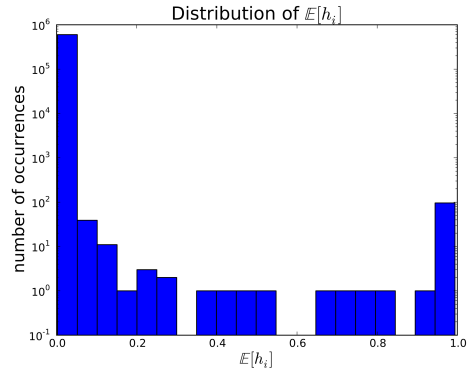

Figure 1. $Q$ imposes a sparse distribution on $h$; $Q(h_i) < .01$ 99.7% of the time in this example histogram of values of $Q(h_i)$ for 6,000 different hidden units from a trained model applied to 100 different $6 \times 6$ image patches.

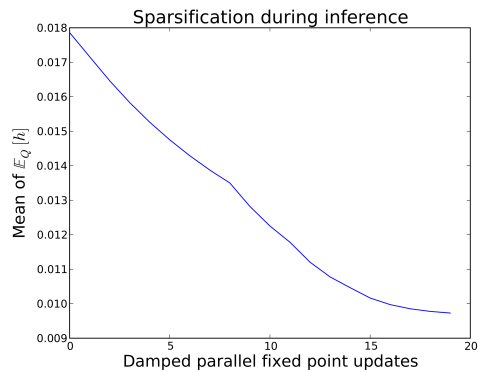

Figure 2. The inference procedure sparsifies the representation due to the explaining-away effect.

richer approximation than the fully factorized family used in the mean field approximation. It allows us to capture the tight correlation between each spike variable and its corresponding slab variable while still allowing simple and efficient inference in the approximating distribution. It also avoids a pathological condition in the mean field distribution where $Q(s_i)$ can never be updated if $Q(h_i) = 0$.

Observing that eq. (4) is an instance of the Euler-Lagrange equation, we find that the solution must take the form

$$Q(h_i) = \hat{h}_i,$$
$$Q(s_i \mid h_i) = \mathcal{N}(s_i \mid h_i \hat{s}_i, \ (\alpha_i + h_i W_i^T \beta W_i)^{-1}) \quad (5)$$

where $\hat{h}_i$ and $\hat{s}_i$ must be found by an iterative process. In a typical application of variational inference, the iterative process consists of sequentially applying fixed point equations that give the optimal value of the parameters $\hat{h}_i$ and $\hat{s}_i$ for one factor $Q(h_i, s_i)$ given the value all of the other factors' parameters. This is for example the approach taken by Titsias and Lázaro-Gredilla (2011) who independently developed a variational inference procedure for the same problem. This process is only guaranteed to decrease the KL divergence if applied to each factor sequentially, i.e. first updating $\hat{h}_1$ and $\hat{s}_1$ to optimize $Q(h_1, s_1)$, then updating $\hat{h}_2$ and $\hat{s}_2$ to optimize $Q(h_2, s_2)$, and so on. In a typical application of variational inference, the optimal values for each update are simply given by the solutions to the Euler-Lagrange equations. For S3C, we make three deviations from this standard approach.

Because we apply S3C to very large-scale problems, we need an algorithm that can fully exploit the benefits of parallel hardware such as GPUs. Sequential updates across all $N$ factors require far too much run-time to be competitive in this regime.

We have considered two different methods that enable parallel updates to all units. In the first method, we start each iteration by partially minimizing the KL divergence with respect to $\hat{s}$. The terms of the KL divergence that depend on $\hat{s}$ make up a quadratic function so this can be minimized via conjugate gradient descent. We implement conjugate gradient descent efficiently by using the R-operator to perform Hessian-vector products rather than computing the entire Hessian explicitly (Schraudolph, 2002). This step is guaranteed to improve the KL divergence on each iteration. We next update $\hat{h}$ in parallel, shrinking the update by a damping coefficient. This approach is not guaranteed to decrease the KL divergence on each iteration but it is a widely applied approach that works well in practice (Koller and Friedman, 2009).

With the second method (Algorithm 1), we find in practice that we obtain faster convergence, reaching equally good solutions by replacing the conjugate gradient update to $\hat{s}$ with a more heuristic approach. We use a parallel damped update on $\hat{s}$ much like what we do for $\hat{h}$. In this case we make an additional heuristic modification to the update rule which is made necessary by the unbounded nature of $\hat{s}$. We clip the update to $\hat{s}$ so that if $\hat{s}_{\text{new}}$ has the opposite sign from $\hat{s}$, its magnitude is at most $\rho\hat{s}$. In all of our experiments we used $\rho = 0.5$ but any value in $[0, 1]$ is sensible. This prevents a case where multiple mutually inhibitory $s$ units inhibit each other so strongly that rather than being driven to 0 they change sign and actually increase in magnitude. This case is a failure mode of the parallel updates that can result in $\hat{s}$ amplifying without bound if clipping is not used.

Figure 1 shows that our E-step produces a sparse representation. Figure 2 shows that the explaining-away

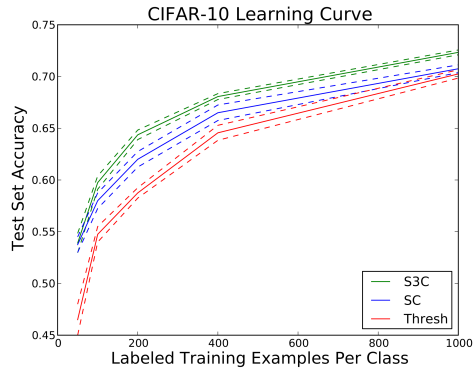

Figure 3. Semi-supervised classification accuracy on subsets of CIFAR-10. Thresholding, the best feature extractor on the full dataset, performs worse than sparse coding when few labels are available. S3C improves upon sparse coding's advantage.

effect incrementally makes the representation more sparse.

---

**Algorithm 1** Fixed-Point Inference

Let K be a user-defined number of inference updates to run (e.g. 20)
Initialize $\hat{h}^{(0)} = \sigma(b)$ and $\hat{s}^{(0)} = \mu$.
**for** k=0:K **do**
  Compute the individually optimal value $\hat{s}_i^*$ for each $i$ simultaneously:
  $$\hat{s}_i^* = \frac{\mu_i \alpha_{ii} + v^T \beta W_i - W_i \beta \left[\sum_{j \neq i} W_j \hat{h}_j \hat{s}_j^{(k)}\right]}{\alpha_{ii} + W_i^T \beta W_i}$$
  Clip reflections by assigning
  $$c_i = \rho \text{sign}(\hat{s}_i^*) |\hat{s}_i^{(k)}|$$
  for all $i$ such that $\text{sign}(\hat{s}_i^*) \neq \text{sign}(\hat{s}_i^{(k)})$ and $|\hat{s}_i^*| > \rho |\hat{s}_i^{(k)}|$, and assigning $c_i = \hat{s}_i^*$ for all other $i$.
  Damp the updates by assigning
  $$\hat{s}_i^{(k+1)} = \eta c + (1-\eta) \hat{s}^{(k)}$$
  where $\eta \in (0, 1]$.
  Compute the individually optimal values for $\hat{h}$:
  $$\hat{h}_i^* = \sigma \left( \left( v - \sum_{j \neq i} W_j \hat{s}_j^{(k+1)} \hat{h}_j^{(k)} - \frac{1}{2} W_i \hat{s}_i^{(k+1)} \right)^T \beta W_i \hat{s}_i^{(k+1)} + b_i \right.$$
  $$\left. - \frac{1}{2} \alpha_{ii} (\hat{s}_i^{(k+1)} - \mu_i)^2 - \frac{1}{2} \log(\alpha_{ii} + W_i^T \beta W_i) + \frac{1}{2} \log(\alpha_{ii}) \right)$$
  Damp the update to $\hat{h}$:
  $$\hat{h}^{(k+1)} = \eta \hat{h}^* + (1-\eta) \hat{h}^{(k)}$$
**end for**

---

## 5. Performance results

Our inference scheme achieves very good computational performance, both in terms of memory consumption and in terms of runtime. The computational bottleneck in our classification pipeline is SVM training, not feature learning or feature extraction.

Comparing the computational cost of our inference scheme to others is a difficult task because it could be confounded by differences in implementation and because it is not clear exactly what sparse coding prob-lem is equivalent to an equivalent spike-and-slab sparse coding problem. However, we observed informally during our supervised learning experiments that feature extraction using S3C took roughly the same amount of time as feature extraction using sparse coding.

In Fig. 4, we show that our improvements to spike-and-slab inference performance allow us to scale spike-and-slab modeling to the problem sizes needed for object recognition tasks.

As a large-scale test of our inference scheme's ability, we trained over 8,000 densely-connected filters on full $32 \times 32$ color images. Some example filters are presented in Fig. 5. This exercise demonstrated that our approach scales well to large (over 3,000 dimensional) inputs, though it is not yet known how to use features for classification as effectively as patch-based features which can be incorporated into a convolutional architecture with pooling. For comparison, to our knowledge the largest image patches used in previous spike-and-slab models with lateral interactions were $16 \times 16$ (Garrigues and Olshausen, 2008).

Finally, we provide empirical justification for our heuristic inference method. Timing experiments presented in Fig. 6 show that the heuristic method is consistently faster than the conjugate gradient method.

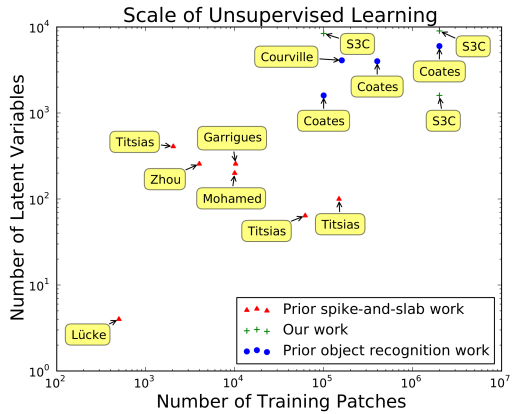

Figure 4. Our inference scheme enables us to extend spike-and-slab modeling from small problems to the scale needed for object recognition. Previous object recognition work is from (Coates and Ng, 2011; Courville et al., 2011a). Previous spike-and-slab work is from (Mohamed et al., 2011; Zhou et al., 2009; Garrigues and Olshausen, 2008; Lücke and Sheikh, 2011; Titsias and Lázaro-Gredilla, 2011).

## 6. Classification results

We conducted experiments to evaluate the usefulness of S3C features for supervised learning on the CIFAR-10 and CIFAR-100 (Krizhevsky and Hinton, 2009) datasets. Both datasets consist of color images of objects such as animals and vehicles. Each contains 50,000 train and 10,000 test examples. CIFAR-10 contains 10 classes while CIFAR-100 contains 100 classes, so there are fewer labeled examples per class in the case of CIFAR-100.

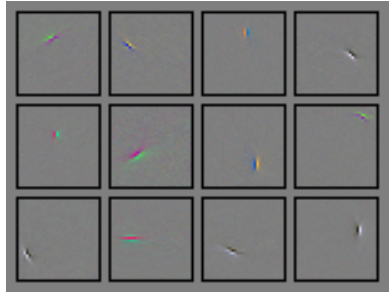

Figure 5. Example filters from a dictionary of over 8,000 learned on full 32x32 images.

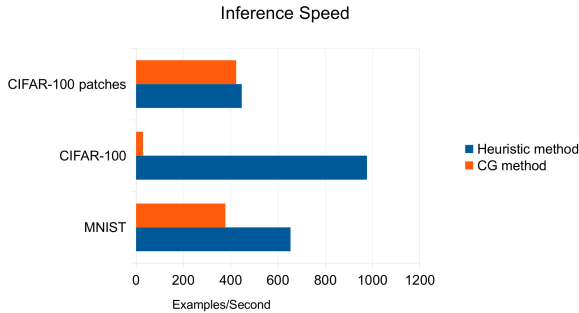

Figure 6. The heuristic method is consistently faster than the conjugate gradient method. The inference speed for each method was computed based on the inference time for the same set of 100 examples from each dataset. In all cases we report the best speed after searching over hyperparameters controlling the amount of damping / conjugate gradient steps to apply at each update.

For all experiments, we used the same overall procedure as Coates and Ng (2011) except for feature learning. CIFAR-10 consists of $32 \times 32$ images. We train our feature extractor on $6 \times 6$ contrast-normalized and ZCA-whitened patches from the training set. At test time, we extract features from all $6 \times 6$ patches on an image, then average-pool them. The average-pooling regions are arranged on a non-overlapping grid. Finally, we train an L2-SVM with a linear kernel on the pooled features.

Coates and Ng (2011) used 1600 basis vectors in all of their sparse coding experiments. They post-processed the sparse coding feature vectors by splitting them into the positive and negative part for a total of 3200 features per average-pooling region. They average-pool on a $2 \times 2$ grid for a total of 12,800 features per image (i.e. each element of the $2 \times 2$ grid averages over a block with sides $\lceil (32 - 6 + 1)/2 \rceil$ or $\lfloor (32 - 6 + 1)/2 \rfloor$). We used $\mathbb{E}_Q[h]$ as our feature vector. This does not have a negative part, so using a $2 \times 2$ grid we would have only 6,400 features. In order to compare with similar sizes of feature vectors we used a $3 \times 3$ pooling grid for a total of 14,400 features (i.e. each element of the $3 \times 3$ grid averages over $9 \times 9$ locations).

| Model | Latent Dimension | Pooling Structure | Best 5-fold CV Accuracy |
|-------|------------------|-------------------|------------------------|
| S3C   | 1600             | $3 \times 3$      | **51.3 %**             |
| SC    | 1600             | $2 \times 2$      | 48.4 %                 |
| SC    | 800              | $3 \times 3$      | 48.7 %                 |
| OMP-1 | 1600             | $2 \times 2$      | 49.1 %                 |
| OMP-1 | 800              | $3 \times 3$      | 47.1 %                 |

Table 1. CIFAR-100 validation accuracy. These results demonstrate that using S3C for the detection layer improves upon OMP-1 (the best detector layer for CIFAR-10) and sparse coding.

### 6.1. CIFAR-10

We use CIFAR-10 to evaluate our hypothesis that S3C resembles a more regularized version of sparse coding.

On the full dataset, S3C achieves a test set accuracy of $78.3 \pm 0.9$ % with 95% confidence. Coates and Ng (2011) do not report test set accuracy for sparse coding with "natural encoding" (i.e., extracting features in a model whose parameters are all the same as in the model used for training) but sparse coding with different parameters for feature extraction than training achieves an accuracy of $78.8 \pm 0.9\%$ (Coates and Ng, 2011). Since we have not enhanced our performance by modifying parameters at feature extraction time these results seem to indicate that S3C is roughly equivalent to sparse coding for this classification task. S3C also outperforms ssRBMs, which require 4,096 basis vectors per patch and a $3 \times 3$ pooling grid to achieve $76.7 \pm 0.9\%$ accuracy. All of these approaches are close to the best result using the pipeline from Coates and Ng (2011) of 81.5% achieved using thresholding of linear features learned with OMP-1. These results show that S3C is a useful feature extractor that performs comparably to the best approaches when large amounts of labeled data are available.

We then tested the regularizing effect of S3C by training the SVM on small subsets of the CIFAR-10 training set, but using features that were learned on patches drawn from the entire CIFAR-10 train set. The results, summarized in Figure 3, show that S3C has the advantage over both thresholding and sparse coding for a wide range of amounts of labeled data. (In the extreme low-data limit, the confidence interval becomes too large to distinguish sparse coding from S3C).

### 6.2. CIFAR-100

Having verified that S3C features help to regularize a classifier, we proceed to use them to improve performance on the CIFAR-100 dataset, which has ten times as many classes and ten times fewer labeled examples per class.

We compare S3C to two other feature extraction methods: OMP-1 with thresholding, which Coates and Ng (2011) found to be the best feature extractor on CIFAR-10, and sparse coding, which is known to perform well when less labeled data is available. We evaluated only a single set of hyperparameters for S3C. For sparse coding and OMP-1 we searched over the same set of hyperparameters as Coates and Ng (2011) did: $\{0.5, 0.75, 1.0, 1.25, 1.25\}$ for the sparse coding penalty and $\{0.1, 0.25, 0.5, 1.0\}$ for the thresholding value. In order to use a comparable amount of computational resources in all cases, we used 1600 hidden units and a $3 \times 3$ pooling grid for S3C, while for the other two methods, which double their number of features via sign splitting, we considered $2 \times 2$ pooling with 1600 latent variables and $3 \times 3$ pooling with 800 latent variables. These results are summarized in Table 1.

The best result to our knowledge on CIFAR-100 is $54.8 \pm 1\%$ (Jia and Huang, 2011), achieved using a learned pooling structure on top of "triangle code" features from a dictionary learned using k-means. This feature extractor is very similar to thresholded OMP-1 features and is known to perform slightly worse on CIFAR-10. Table 1 shows that S3C is the best known detector layer on CIFAR-100. If combined with the pooling strategy of Jia and Huang (2011) it has the potential to improve on the state of the art. Using a pooling strategy of concatenating $1 \times 1$, $2 \times 2$ and $3 \times 3$ pooled features we achieve a test set accuracy of $53.7 \pm 1\%$.

## 7. Transfer Learning Challenge

For the NIPS 2011 Workshop on Challenges in Learning Hierarchical Models (Le et al., 2011), the organizers proposed a transfer learning competition. This competition used a dataset consisting of $32 \times 32$ color images, including 100,000 unlabeled examples, 50,000 labeled examples of 100 object classes not present in the test set, and 120 labeled examples of 10 object classes present in the test set. We applied the same approach as on the CIFAR datasets and won the com-petition, with a test set accuracy of 48.6 %. This approach disregards the 50,000 labels and treats this transfer learning problem as a semi-supervised learning problem. We experimented with some transfer learning techniques but the transfer-free approach performed best on leave-one-out cross-validation on the 120 example training set, so we chose to enter the transfer-free technique in the challenge.

## 8. Conclusion

We have motivated the use of the S3C model for unsupervised feature discovery. We have described a variational approximation scheme that makes it feasible to perform learning and inference in large-scale S3C models. We have demonstrated that S3C is an effective feature discovery algorithm for both supervised and semi-supervised learning with small amounts of labeled data. This work addresses two scaling problems: the computation problem of scaling spike-and-slab sparse coding to the problem sizes used in object recognition, and the problem of scaling object recognition techniques to work with more classes.